\documentclass[runningheads]{llncs}
\usepackage{graphicx}
\usepackage{subcaption}
\usepackage{booktabs}
\usepackage{multirow}
\usepackage{comment}
\begin{document}
\title{Combining Fine- and Coarse-Grained Classifiers for Diabetic Retinopathy Detection 
\thanks{Partially funded by Higher Education Commission (HEC) of Pakistan through Faculty Development Program of  National University of Science and Technology (NUST), Pakistan}
}
%
%

\author{Muhammad Naseer Bajwa\orcidID{0000-0002-4821-1056}\inst{1,2} \and
Yoshinobu Taniguchi\orcidID{0000-0002-0625-9123}\inst{3} \and
Muhammad Imran Malik\inst{4} \and
Wolfgang Neumeier\inst{5} \and
Andreas Dengel\inst{1,2} \and
Sheraz Ahmed\orcidID{0000-0002-4239-6520}\inst{2}
}
\authorrunning{M. N. Bajwa et al.}
%

\institute{Technische Universit\"at Kaiserslautern, Kaiserslautern 67663, Germany \and
German Research Center for Artificial Intelligence GmbH (DFKI), Kaiserslautern 67663, Germany \\
\email{bajwa@dfki.uni-kl.de}
\email{\{andreas.dengel,sheraz.ahmed\}@dfki.de}
\and
Osaka Prefecture University, Naka, Sakai, Osaka 599-8531, Japan\\
\email{taniguchi@m.cs.osakafu-u.ac.jp}
\and
School of Electrical Engineering and Computer Science, National University of Science and Technology (NUST), Islamabad 46000, Pakistan\\
\email{malik.imran@seecs.edu.pk}
\and Opthalmology Clinic, Rittersberg 9, 67657 Kaiserslautern, Germany\\
\email{dr.neumeier-kl@web.de}}
\maketitle              
\vspace{-0.7 cm}
\begin{abstract}
Visual artefacts of early diabetic retinopathy in retinal fundus images are usually small in size, inconspicuous, and scattered all over retina. Detecting diabetic retinopathy requires physicians to look at the whole image and fixate on some specific regions to locate potential biomarkers of the disease. Therefore, getting inspiration from ophthalmologist, we propose to combine coarse-grained classifiers that detect discriminating features from the whole images, with a recent breed of fine-grained classifiers that discover and pay particular attention to pathologically significant regions. To evaluate the performance of this proposed ensemble, we used publicly available EyePACS and Messidor datasets. Extensive experimentation for binary, ternary and quaternary classification shows that this ensemble largely outperforms individual image classifiers as well as most of the published works in most training setups for diabetic retinopathy detection. Furthermore, the performance of find-grained classifiers is found notably superior than coarse-grained image classifiers encouraging the development of task-oriented find-grained classifiers modeled after specialist ophthalmologists.

\end{abstract}
\section{Introduction}
Diabetic patients are at constant risk of developing Diabetic Retinopathy (DR) that may eventually lead to permanent vision loss if left unnoticed or untreated. In such patients, increased blood sugar, blood pressure, and cholesterol can cause small blood vessels in retina to protrude and, in due course, haemorrhage blood into retinal layers and/or vitreous humour \cite{amin2017method}. In severe conditions, scar tissues and newly proliferated fragile blood vessels blanket the retina and obstruct incoming light from falling on it. As a result, retina is unable to translate light into neural signals which results in blindness. Diabetic retinopathy advances slowly and gradually and may take years to reach proliferative stage, however, almost every diabetic patient is potentially susceptible to this complication.

Timely diagnosis is the key to appropriate prognosis. Ophthalmologists usually detect DR by examining retinal fundus and looking for any signs of microaneurysms (bulging of blood vessels), blood leakage, and/or neovascularization \cite{akram2014detection}. While the indications of advanced stages of DR are rather prominent, these symptoms remain largely discrete in early stages. Figure \ref{fig:drStages} shows progress of DR from healthy to proliferative stage in Retinal Fundus Images (RFIs) taken from EyePACS dataset\footnote{https://www.kaggle.com/c/diabetic-retinopathy-detection/data}. It can be observed from the figure that the difference between healthy and early stages of DR are very subtle and not readily discernible. Manual analysis of these images requires highly qualified and specialized ophthalmologists who may not be easily accessible in developing countries or remote areas of developed countries. Even when medical experts are available, large scale analysis of RFIs is highly time-consuming, labour-intensive and prone to human error and bias. Furthermore, manual diagnosis by clinicians is largely subjective and rarely reproducible and, therefore, inter-expert agreement for a certain diagnosis is generally very poor.

Computer-Aided Diagnosis (CAD) based on deep learning can provide easily accessible, efficient and economical solution for large-scale initial screening of many diseases including diabetic retinopathy. CAD can perform objective analysis of the given image and predict standardized and reproducible diagnosis, which is free from any bias or tiredness. It can not only help physicians by reducing their workload but can also outreach to underprivileged population and afford them the opportunity of swift and cost-effective initial screening, which may effectively prevent advancement of disease into severer stage.
\begin{figure}[h]
\centering
\addtocounter{figure}{-1}
    \begin{subfigure}{0.19\textwidth}
        \includegraphics[width=\textwidth]{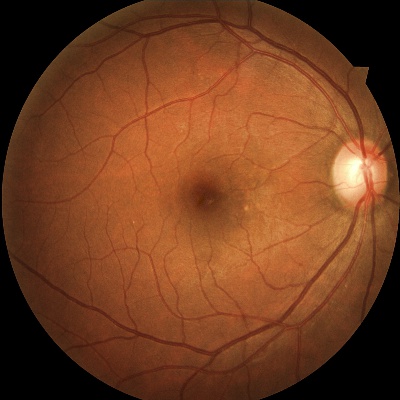}
        \caption{(a) Healthy}
        \label{fig:healthy}
    \end{subfigure}
    \begin{subfigure}{0.19\textwidth}
        \includegraphics[width=\textwidth]{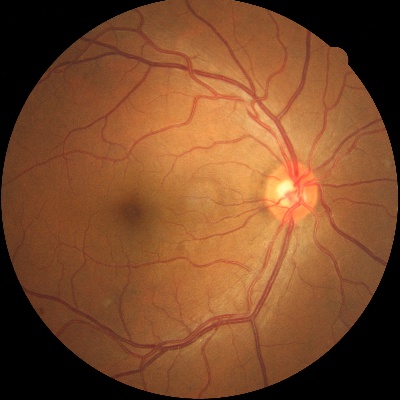}
        \caption{(b) Mild}
        \label{fig:mild}
    \end{subfigure}
    \begin{subfigure}{0.19\textwidth}
        \includegraphics[width=\textwidth]{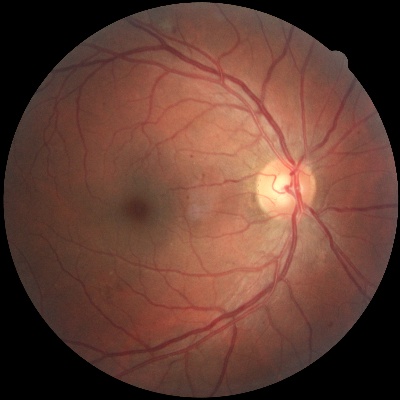}
        \caption{(c) Moderate}
        \label{fig:moderate}
    \end{subfigure}
    \begin{subfigure}{0.19\textwidth}
        \includegraphics[width=\textwidth]{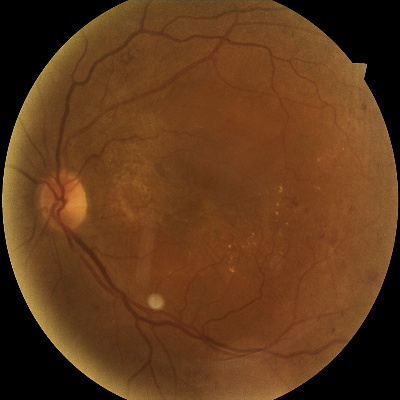}
        \caption{(d) Severe}
        \label{fig:severe}
    \end{subfigure}
    \begin{subfigure}{0.19\textwidth}
        \includegraphics[width=\textwidth]{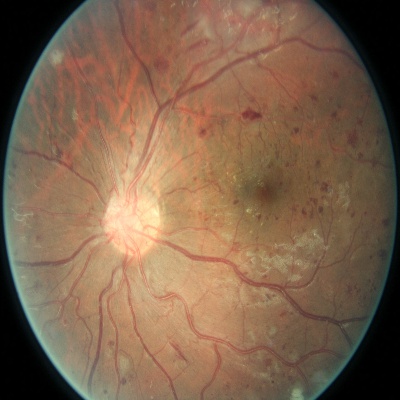}
        \caption{(e) Proliferative}
        \label{fig:prolif}
    \end{subfigure}
    \caption{Progression of diabetic retinopathy from healthy to proliferative stage is subtle and gradual. Images are taken from EyePACS train set.}
    \label{fig:drStages}
\end{figure}
Convolutional Neural Networks (CNNs) are computer algorithms inspired by biological visual cortex. They work especially well in visual recognition tasks. CNNs have been used to perform at par with or even outperform humans in various challenging image recognition problems \cite{DBLP:journals/corr/abs-1712-00559,lee2016generalizing}. Today automated image recognition can be divided into coarse-grained classification and fine-grained classification. In former case, images are classified into high-level categories like humans, animals, vehicles and other objects in a natural scene, for example. In later case, classification is focused on low-level categories like species of dogs or models of cars etc. Fine-grained classification is particularly challenging owning to high intra-class variations and low inter-class variations. Although DR is also a fine-grained classification task, it has normally been addressed using simple coarse-grained classification algorithms.

In this work we used a combination of general and fine-grained deep CNNs to analyze RFIs and predict automated diagnosis for DR. We used two of the most popular conventional image classification architectures i.e Residual Networks \cite{he2016resnet} and Densely Connected Networks \cite{huang2017densenet}, a network search framework called NASNet \cite{zoph2018nasnet} and two recently proposed methods for fine-grained classification namely NTS-Net \cite{yang2018ntsNet} and SBS Layer \cite{recasens2018saliency}. We tried to harvest the combined potential of these two approaches by training them separately and taking their ensemble during inference. We used EyePACS and Messidor datasets for evaluation. Since previous researches have used vastly disparate experimental setups, we cannot directly compare our results with most of them. However, we performed a broad range of experiments, following the most common problem settings in the literature like normal vs abnormal, referable vs non-referable, ternary and quaternary classification in order to define benchmarks which will afford future works with an opportunity of fair comparison.

\subsection{Related Work}\label{sec:relatedWork}
Over the past decade, machine learning and deep learning have been used to detect various pathologies, segment vessels and classify DR grades using RFIs. Welikala et al. \cite{welikala2014automated}~ detected proliferative DR by identifying neovascularization. They used an ensemble of two networks trained separately on 100 different patches for each network. The patches are taken from a selected set of 60 images collected from Messidor \cite{messidor2014feedback} and a private dataset. Since the dataset had only 60 images they performed leave-one-out cross validation and achieved 0.9505 Area Under the Curve (AUC) and sensitivity of 1 with specificity of 0.95 at the optimal operating point. Wang et al. \cite{wang2017zoom} identified suspicious regions in RFIs and classified DR into normal (nDR) vs abnormal (aDR) and referable (rDR) vs non-referable (nrDR). They developed a CNN based model called Zoom-in-Network to identify important regions. To classify an image the network uses the overview of the whole images and pays particular attention to important regions. They took 182 images from EyePACS dataset and let a trained ophthalmologist draw bounding boxes around 306 lesions. On Messidor dataset they achieved 0.921 AUC, 0.905 accuracy and 0.960 sensitivity at 0.50 specificity for nDR vs aDR.

Gulshan et al. \cite{gulshan2016development} conducted a comprehensive study to distinguish rDR from nrDR grades. They trained a deep CNN on 128175 fundus images from a private dataset and tested on 9963 images from EyePACS-I and 1748 images of Messidor-2. They achieved AUC of 0.991 on EyePACS-I and 0.990 on Messidor-2. Guan et al. \cite{guan2018said} proposed that modeling each classifier after individual human grader instead of training a single classifier using average grading of all human experts improves classification performance. They trained 31 classifiers using a dataset of total 126522 images collected from EyePACS and three other clinics. The method is tested on 3547 images from EyePACS-I and Messidor-2, and achieved 0.9728 AUC, 0.9025 accuracy, and 0.8181 specificity at 0.97 sensitivity. However, it would have been more interesting if they had provided comparison of their suggested approach with ensemble of 31 networks modeled after average grading. Costa et al. \cite{costa2018end} used adversarial learning to synthesize color retinal images. However, the performance of their classifier trained on synthetic images was less than the classifier trained on real images. Aujih et al. \cite{aujih2018analysis} found that blood vessels play important role in disease classification and fundus images without blood vessels resulted in poor performance by the classifier.

The role of multiple filter sizes in learning fine-grained features was studied by Vo et al. \cite{vo2016new}. To this end they used VGG network with extra kernels and combined kernels with multiple loss networks. They achieved 0.891 AUC for rDR vs nrDR and 0.870 AUC for normal vs abnormal on Messidor dataset using 10-fold cross validation. Somkuwar et al. \cite{somkuwar2015intensity} performed classification of hard exudates by exploiting intensity features using 90 images from Messidor dataset and achieved 100\% accuracy on normal and 90\% accuracy on abnormal images. Seoud et al. \cite{seoud2016red} focused on red lesions in RFIs, like haemorrhages and microaneurysms, and detected these biomarkers using dynamic shape features in order to classify DR. They achieved 0.899 AUC and 0.916 AUC for nDR vs aDR and rDR vs nrDR, respectively on Messidor. Rakhlin et al. \cite{rakhlin2018diabetic} used around 82000 images taken from EyePACS for training and around 7000 EyePACS images and 1748 images from Messidor-2  for testing their deep learning based classifier. They achieved 0.967 AUC on Messidor and 0.923 AUC on EyePACS for binary classification. Ramachandran et al. \cite{ramachandran2018diabetic} used 485 private images and 1200 Messidor images to test a third party deep learning based classification platform, which was trained on more than 100000 images. Their validation gave them 0.980 AUC on Messidor dataset for rDR vs nrDR classification. Quellec et al. \cite{quellec2017deep} capitalized a huge private dataset of around 110000 images and around 89000 EyePACS images to train and test a classifier for rDR vs nrDR grades and achieved 0.995 AUC on EyePACS.
\vspace{-0.4 cm}
\section{Materials and Methods}
\vspace{-0.3 cm}
This section provides details on the datasets used in this work and the ensemble methodology employed to perform classification.
\subsection{Datasets}
We used EyePACS dataset published publicly by Kaggle for a competition on Diabetic Retinopathy Detection. Table \ref{tab:EyePACS} gives overview of EyePACS dataset. Although this dataset is very large in size, only about 75\% of its images are of reasonable quality that they can be graded by human experts \cite{rakhlin2018diabetic}. EyePACS is graded on a scale of 0 to 4 in accordance with International Clinical Diabetic Retinopathy (ICDR) guidelines \cite{ICDR2002}. However, low gradability of this dataset raises suspicions on the fidelity of labels provided with each image. We pruned the train set to get rid of 657 completely uninterpretable images. For testing on EyePACS we used 33423 images randomly taken from test set.

\begin{table}[h]
\caption{Overview of EyePACS Dataset. IRMA stands for IntraRetinal Microvascular Abnormalities}\label{tab:EyePACS}
\centering
\resizebox{\textwidth}{!}{%
\begin{tabular}{@{}cp{6cm}cccc@{}}
\toprule
\multirow{2}{*}{Severity Grade} & \multicolumn{1}{c}{\multirow{2}{*}{Criterion}}                                                                                                                                                                                   & \multicolumn{2}{c}{Train Set} & \multicolumn{2}{c}{Test Set} \\ \cmidrule(l){3-6} 
                                & \multicolumn{1}{c}{}                                                                                                                                                                                                             & Images      & Percentage      & Images      & Percentage     \\ \midrule
0                               & No Abnormalities                                                                                                                                                                                                                 & 25810       & 73.48           & 39533       & 73.79          \\ \hline
1                               & Microaneurysms Only                                                                                                                                                                                                              & 2443        & 6.95            & 3762        & 7.02           \\ \hline
2                               & More than just microaneurysms but less than Grade 3                                                                                                                                                                              & 5292        & 15.07           & 7861        & 14.67          \\ \hline
3                               & \begin{tabular}[c]{@{}p{6cm}@{}}More than 20 intraretinal hemorrhages in each of 4 quadrants\\ OR Definite venous beading in 2+ quadrants\\ OR Prominent IRMA in 1+ quadrant\\ AND no signs of proliferative retinopathy\end{tabular} & 873         & 2.48            & 1214        & 2.27           \\ \hline
4                               & \begin{tabular}[c]{@{}p{6cm}@{}}Neovascularization\\ OR Vitreous/preretinal hemorrhage\end{tabular}                                                                                                                                   & 708         & 2.02            & 1206        & 2.25           \\ \hline
\multicolumn{2}{c}{Total}                                                                                                                                                                                                                                          & 35126       & 100             & 53576       & 100            \\ \bottomrule
\end{tabular}
}
\end{table}

Messidor dataset consists of 1200 images collected at three different clinics in France. Each clinic contributed 400 images. This dataset is graded for DR on a scale of 0 to 3 following the criteria given in Table \ref{tab:Messidor}. Messidor dataset is validated by experts and is, therefore, of higher quality than EyePACS in terms of both image quality and labels.

\begin{table}[h]
\caption{Overview of Messidor Dataset}\label{tab:Messidor}
\centering
\begin{tabular}{clcc}
\hline
Severity Grade & \multicolumn{1}{c}{Criterion}                                                                                                                          & Images & Percentage \\ \hline
0              & \begin{tabular}[c]{@{}l@{}}No microaneurysms\\ AND No haemorrhages\end{tabular}                                                                        & 546    & 45.50      \\ \hline
1              & \begin{tabular}[c]{@{}l@{}}Microaneurysms \textless{}= 5\\ AND No haemorrhages\end{tabular}                                                            & 153    & 12.75      \\ \hline
2              & \begin{tabular}[c]{@{}l@{}}5 \textless Microaneurysms \textless 15\\ AND 0 \textless Haemorrhages \textless 5\\ AND No Neovascularization\end{tabular} & 247    & 20.58      \\ \hline
3              & \begin{tabular}[c]{@{}l@{}}Microaneurysms \textgreater{}= 15\\ OR Haemorrhages \textgreater{}= 5\\ OR Neovascularization\end{tabular}                  & 254    & 21.17      \\ \hline
\multicolumn{2}{c}{Total}                                                                                                                                               & 1200   & 100        \\ \hline
\end{tabular}%
\end{table}

\subsection{Methodology}
\begin{figure}[b!]
\centering
\includegraphics[width=\textwidth]{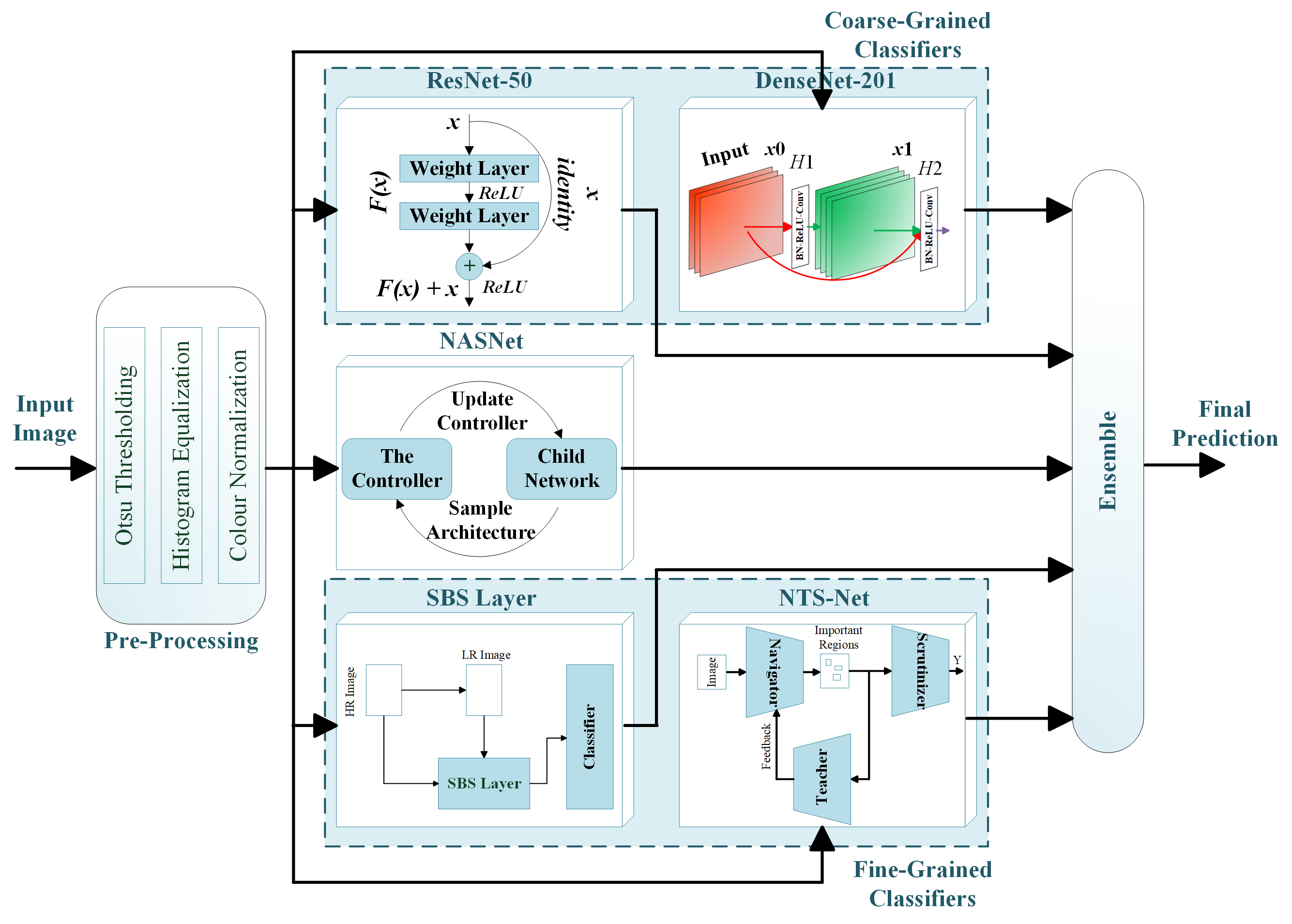}
\caption{System Overview: Combining Coarse-Grained and Fine-Grained Classifiers}
\label{fig:systemOverview}
\end{figure}
Figure \ref{fig:systemOverview} illustrates complete pipeline of the system combining coarse-grained and find-grained classifiers. 
Before feeding an image to the network, we first applied Otsu Thresholding to extract and crop retinal rim from RFI and get rid of irrelevant black background. Since the images in both datasets are taken with different cameras and under different clinical settings, they suffer from large brightness and color variations. We used adaptive histogram equalization to normalize brightness and enhance the contrast of visual artefacts which are critical for DR detection. Since the images are in RGB colour space, we first translate them into YCbCr colour space to distribute all luminosity information in Y channel and colour information in Cb and Cr channels. Adaptive histogram equalization is then applied on Y channel only and the resultant image is converted back to RGB colour space. We further normalized the images by subtracting local average colour from each pixel to highlight the foreground and help our network detect small features. Figure \ref{fig:PreProcess} shows the effects of preprocessing steps on RFIs. These pre-processed images are then used to train all five networks individually. During inference, each network gives diagnosis which are ensembled to calculate the final prediction.
\begin{figure}[t]
    \centering
    \addtocounter{figure}{-1}
    \begin{subfigure}{0.32\textwidth}
        \includegraphics[width=\textwidth]{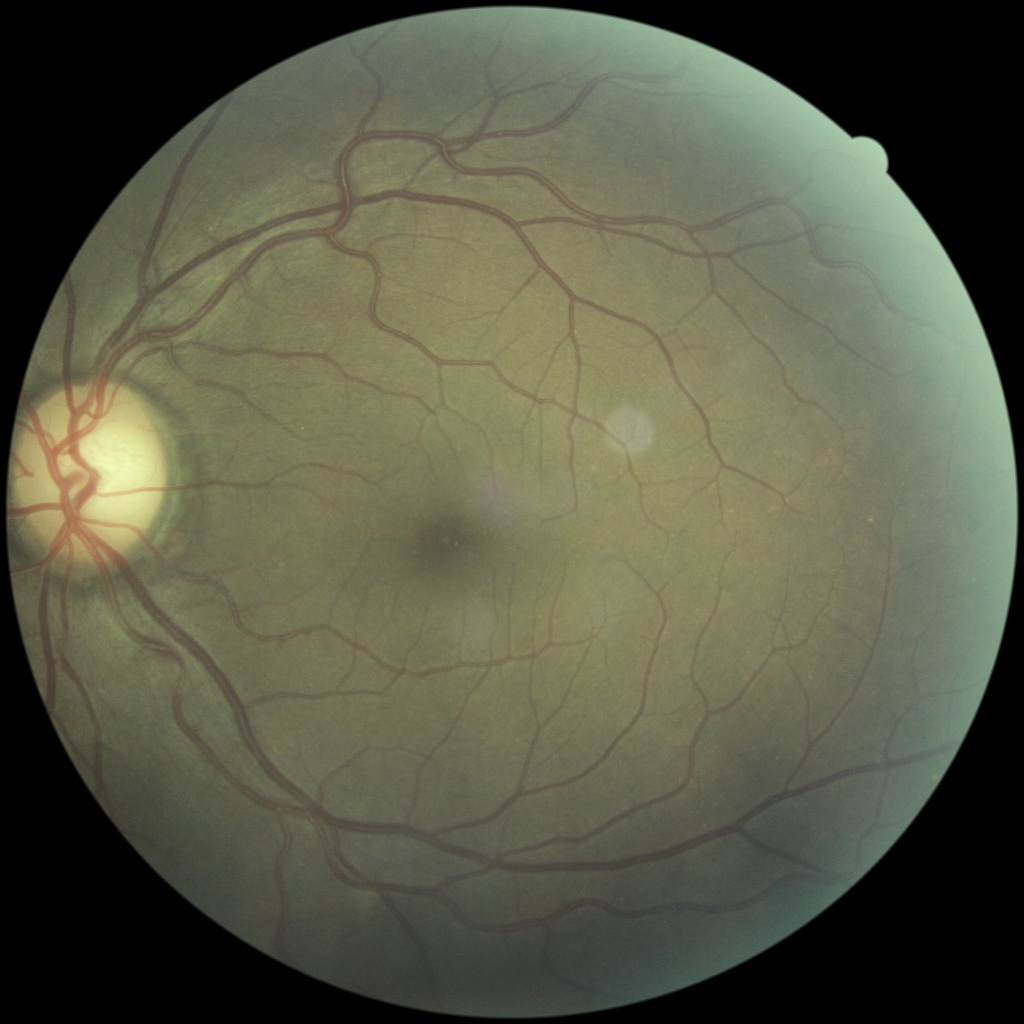}
        \caption{(a) Original Image before Preprocessing}
        \label{fig:original}
    \end{subfigure}
    \begin{subfigure}{0.32\textwidth}
        \includegraphics[width=\textwidth]{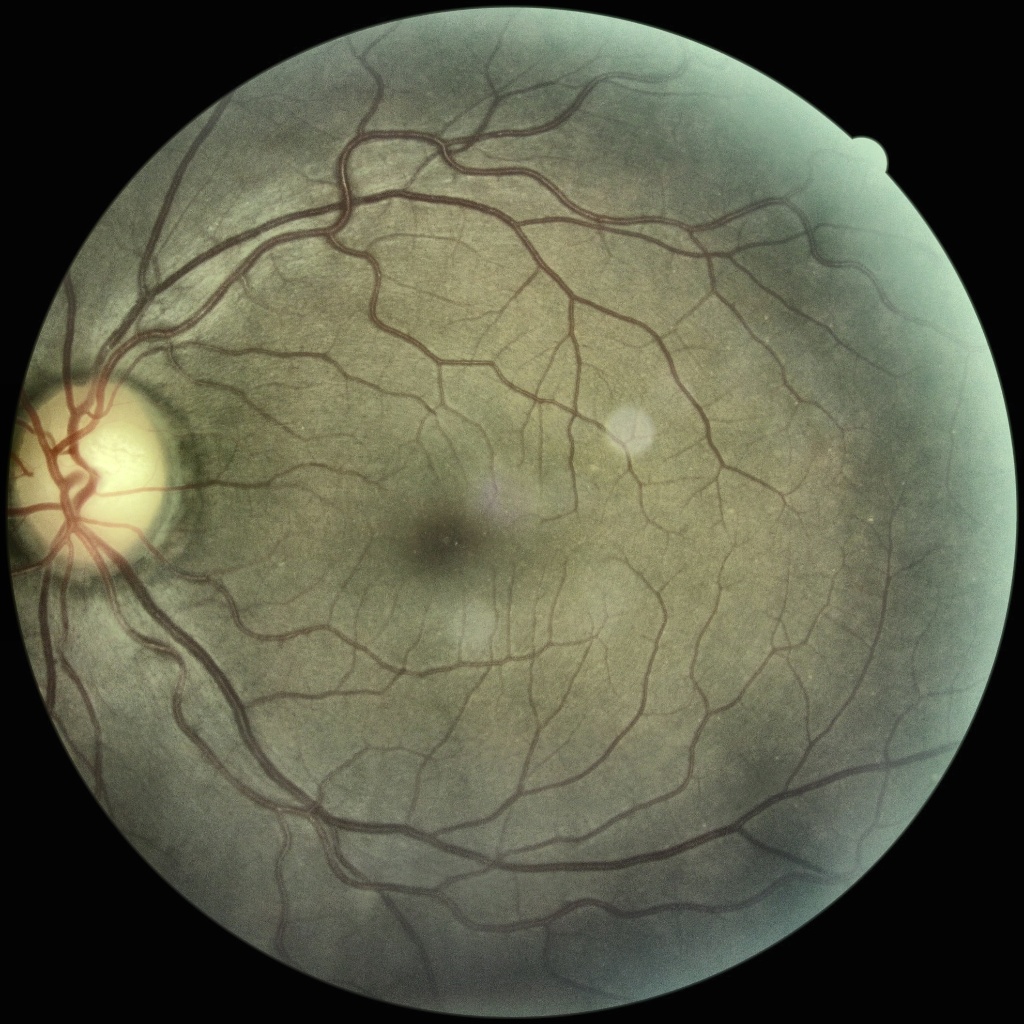}
        \caption{(b) After Contrast Enhancement}
        \label{fig:histEq}
    \end{subfigure}
    \begin{subfigure}{0.32\textwidth}
        \includegraphics[width=\textwidth]{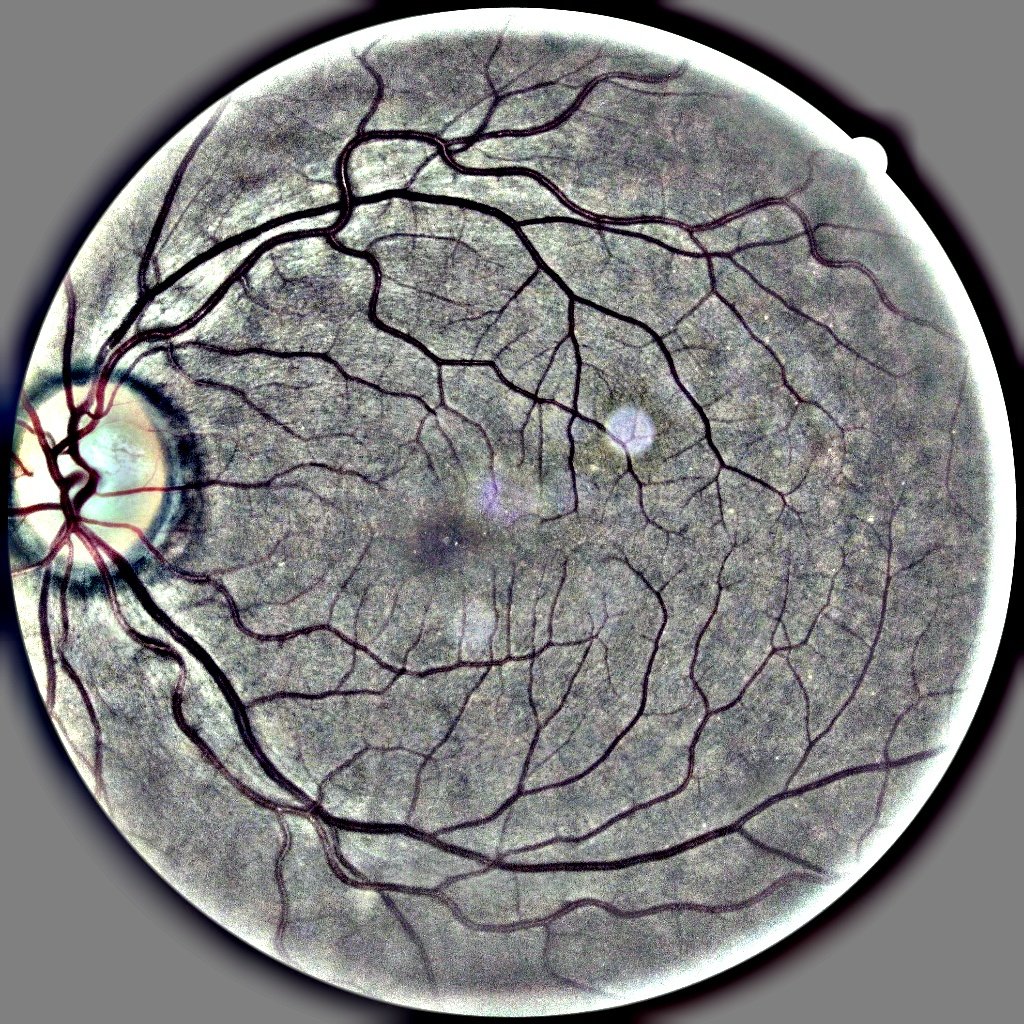}
        \caption{(c) After Local Average Colour Subtraction}
        \label{fig:lvcSubtr}
    \end{subfigure}
    \caption{Effects of preprocessing steps on retinal fundus images}
    \label{fig:PreProcess}
\end{figure}

\subsubsection{Experimental Setup}
From EyePACS train set, we randomly selected 30000 images for training and rest of the 4469 images were used for validation. Test set of EyePACS was used for reporting results on this dataset. From Messidor, we used 800 images for training and 400 images from Lariboisi\`ere Hospital for testing (as done  by Lam et al. \cite{carson2018automated}). We employed a broad range of hyper parameters during training. All networks are initialized with pre-trained weights and fine-tuned on ophthalmology datasets. To evaluate these models on EyePACS and Messidor datasets under similar problem settings, we first parallelized DR grades of both datasets using criteria given in Figure \ref{fig:grades}.

\begin{figure}[b]
\centering
\includegraphics[width=0.70\textwidth]{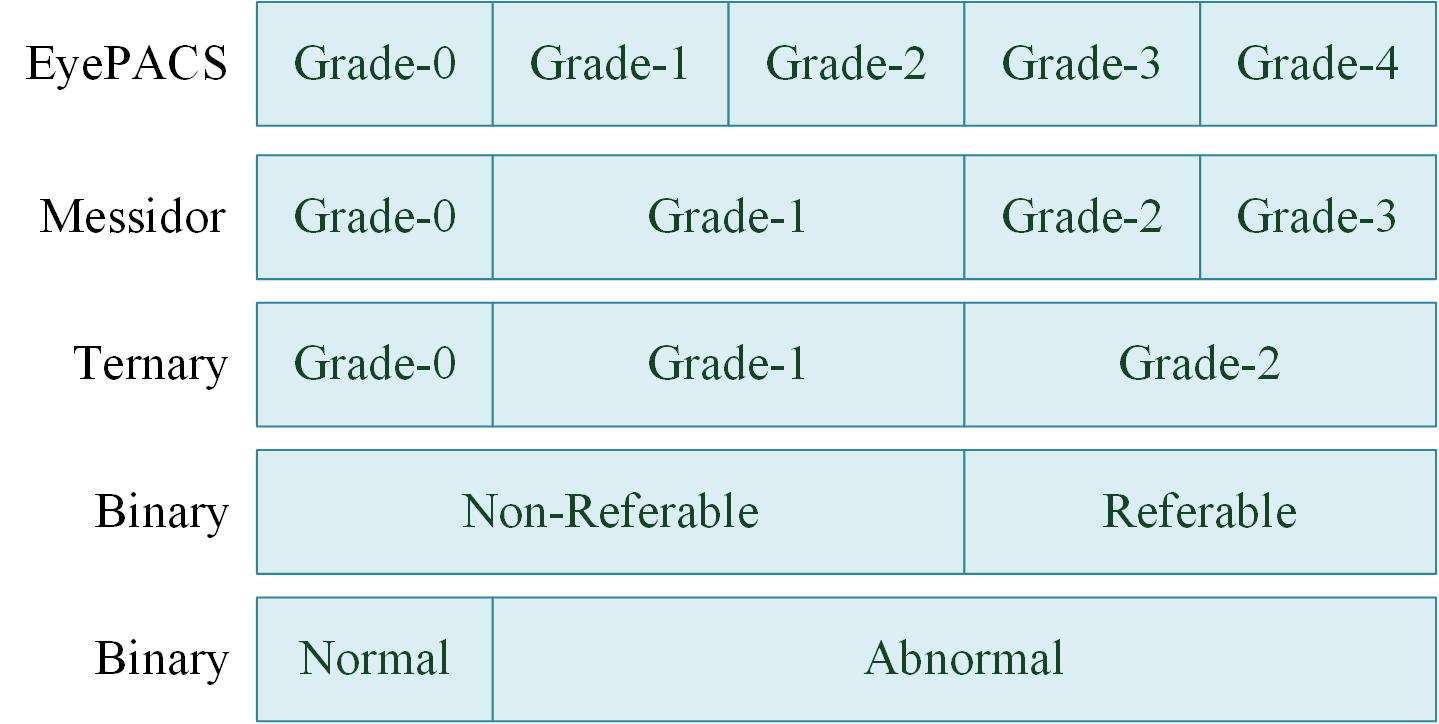}
\caption{Conversion of EyePACS grades to Quaternary, Ternary and Binary Classification}
\label{fig:grades}
\end{figure}

\section{Results and Analysis}
From section \ref{sec:relatedWork} we observe that previous works on EyePACS and Messidor have used disparate train and test splits and different classification tasks for example Quaternary, Ternary and Binary (rDR vs nrDR and nDR vs aDR). Furthermore, different researchers use different performance metrics to evaluate their method. Therefore, in such scenario comparison of any two works in not directly possible \cite{abramoff2016improved}. However, we conducted extensive experiments to perform all four classification tasks mentioned above and report comprehensive results to allow a rough comparison with some of the published state-of-the-art results on these datasets. 

\subsection{Results of Binary Classification}
As discussed above, many previous works focus primarily on binary classification as nDR vs aDR or rDR vs nrDR grading. The criteria to convert 4 or 5 grades into binary grades is given in Figure \ref{fig:grades}. For our binary classification, the number of images used for training, validation and testing from EyePACS and Messidor are given in Table \ref{tab:distNAB} and \ref{tab:distRNR}. It can be seen from the tables that there is extensive class imbalance between both classes. Table \ref{tab:allResults} provides detailed performance metrics for all classification tasks including nDR vs aDR classification. Our results are competitive to that of Wang et al. in terms of accuracy and for all other metrics we outperform them. It should be noted here that Wang et al. performed 10-fold cross validation and although their sensitivity of 96 is higher than our 89.75, it is calculated at 50\% specificity while ours is at 90\% specificity.

\begin{table}
\resizebox{\textwidth}{!}{%
\parbox[t]{.49\linewidth}{
\centering
\caption{Class Distribution for Normal vs Abnormal Classification}\label{tab:distNAB}
\resizebox{0.50\textwidth}{!}{ 
\begin{tabular}{lcccccc}
\hline
\multicolumn{1}{c}{\multirow{2}{*}{Grade}} & \multicolumn{3}{c}{EyePACS} & \multicolumn{3}{c}{Messidor} \\ \cline{2-7} 
\multicolumn{1}{c}{}                       & Train  & Validate  & Test   & Train   & Validate   & Test  \\ \hline
Normal                                     & 22668  & 2744      & 24741  & 346     & 49         & 151   \\
Abnormal                                   & 7332   & 1725      & 8682   & 354     & 51         & 249   \\
Total                                      & 30000  & 4469      & 33423  & 700     & 100        & 400   \\ \hline
\end{tabular}
}
}
\hfill
\parbox[t]{.49\linewidth}{
\centering
\caption{Class Distribution for Referable vs Non-Referable Classification}\label{tab:distRNR}
\resizebox{0.50\textwidth}{!}{ 
\begin{tabular}{lcccccc}
\hline
\multicolumn{1}{c}{\multirow{2}{*}{Grade}} & \multicolumn{3}{c}{EyePACS} & \multicolumn{3}{c}{Messidor} \\ \cline{2-7} 
\multicolumn{1}{c}{}                       & Train  & Validate  & Test   & Train   & Validate   & Test  \\ \hline
Non-Referable                              & 28825  & 4177      & 31937  & 453     & 65         & 181   \\
Referable                                  & 1175   & 292       & 1486   & 247     & 35         & 219   \\
Total                                      & 30000  & 4469      & 33423  & 700     & 100        & 400   \\ \hline
\end{tabular}
}
}
}
\end{table}

\begin{table}[]
\centering
\caption{Detailed Performance Metrics for Various Classification Settings}\label{tab:allResults}
\resizebox{\textwidth}{!}{%
\begin{tabular}{@{}lcccccccc@{}}
\toprule
\multicolumn{9}{c}{\textbf{Results of Binary (Normal vs Abnormal) Classification}}                                                                                                                                                       \\ \midrule
\multicolumn{1}{c}{\multirow{2}{*}{\textbf{Model}}} & \multicolumn{2}{c}{\textbf{Accuracy (\%)}} & \multicolumn{2}{c}{\textbf{AUC (\%)}} & \multicolumn{2}{c}{\textbf{Sensitivity (\%)}} & \multicolumn{2}{c}{\textbf{Specificity (\%)}} \\ \cmidrule(l){2-9} 
\multicolumn{1}{c}{}                                & EyePACS              & Messidor            & EyePACS           & Messidor          & EyePACS              & Messidor               & EyePACS              & Messidor               \\ \midrule
NTS-Net                                             & \textbf{88.19}       & 88.00               & 92.72             & 95.20             & 88                   & 88.00                  & 72                   & 87.51                  \\
SBS Layer                                           & 80.11                & 89.50               & 86.20             & 95.17             & 80                   & 89.50                  & 54                   & \textbf{92.07}         \\
ResNet-50                                           & 82.86                & 87.75               & 89.46             & 95.06             & 83                   & 87.75                  & 75          & 90.49                  \\
DenseNet-201                                        & 82.66                & 87.75               & 89.69             & 95.89             & 83                   & 87.75                  & \textbf{77}          & 88.14                  \\
NASNet                                              & 82.19                & 87.25               & 88.49             & 95.04             & 82                   & 87.25                  & 73                   & 89.14                  \\
\textbf{Ensemble}                                            & 87.74                & \textbf{89.75}      & \textbf{93.44}    & \textbf{96.50}    & \textbf{88}          & \textbf{89.75}         & 75                   & 91.44                  \\
Vo et. al                                           & N/A                  & 87.10               & N/A               & 87.00             & N/A                  & 88.2                   & N/A                  & 85.7                   \\
Wang et. al                                         & N/A                  & 90.50               & N/A               & 92.10             & N/A                  & 96                     & N/A                  & 50                     \\
Soud et. al                                         & N/A                  & N/A                 & N/A               & 89.90             & N/A                  & N/A                    & N/A                  & N/A                    \\ \midrule
\multicolumn{9}{c}{\textbf{Results of Binary (Referable vs Non-Referable) Classification}}                                                                                                                                               \\ \midrule
NTS-Net                                             & 94.93                & \textbf{93.25}      & 99.10             & \textbf{96.56}    & 95                   & \textbf{93}            & 75                   & \textbf{94}            \\
SBS Layer                                           & \textbf{95.89}       & 88.75               & \textbf{99.44}    & 94.90             & \textbf{96}          & 89                     & 67                   & 90                     \\
ResNet-50                                           & 95.08                & 86.75               & 98.97             & 94.95             & 95                   & 87                     & 81                   & 89                     \\
DenseNet-201                                        & 94.70                & 89.25               & 99.05             & 95.33             & 95                   & 89                     & 82                   & 91                     \\
NASNet                                              & 91.98                & 87.50               & 97.45             & 95.16             & 92                   & 88                     & \textbf{85}          & 89                     \\
\textbf{Ensemble}                                   & 95.34                & 89.25               & 99.23             & 96.45             & 95                   & 89                     & 81                   & 91                     \\
Lam et. al                                          & N/A                  & 74.5                & N/A               & N/A               & N/A                  & N/A                    & N/A                  & N/A                    \\
Vo et. al                                           & N/A                  & 89.70               & N/A               & 89.10             & N/A                  & 89.3                   & N/A                  & 90                    \\
Wang et. al                                         & N/A                  & 91.10               & N/A               & 95.70             & N/A                  & 97.8                   & N/A                  & 50                    \\
Seoud et. al                                        & N/A                  & 74.5                & N/A               & 91.60             & N/A                  & N/A                    & N/A                  & N/A                    \\
\midrule
\multicolumn{9}{c}{\textbf{Results of Ternary Classification}}                                                                                                                                                                           \\ \midrule
NTS-Net                                             & 84.43                & 84.50               & 94.89             & 94.61             & 84                   & 85                     & 72                   & \textbf{94}            \\
SBS Layer                                           & 76.93                & 84.50               & 90.95             & 94.12             & 77                   & 85                     & 50                   & 91                     \\
ResNet-50                                           & 81.23                & 80.50               & 93.51             & 93.79             & 81                   & 81                     & 74                   & 92                     \\
DenseNet-201                                        & 79.20                & 80.25               & 92.87             & 94.25             & 79                   & 80                     & \textbf{77}          & 93                     \\
NASNet                                              & 78.95                & 81.75               & 91.93             & 94.00             & 79                   & 82                     & 71                   & 89                     \\
\textbf{Ensemble}                                   & \textbf{84.94}       & \textbf{85.25}      & \textbf{95.28}    & \textbf{95.40}    & \textbf{85}          & \textbf{85}            & 73                   & 92                     \\
Lam et. al                                          & N/A                  & 68.8                & N/A               & N/A               & N/A                  & N/A                    & N/A                  & N/A                    \\ \midrule
\multicolumn{9}{c}{\textbf{Results of Quaternary Classification}}                                                                                                                                                                        \\ \midrule
NTS-Net                                             & 82.53                & 74.50               & 95.72             & 91.84             & 83                   & 75                     & \textbf{76}          & \textbf{92}            \\
SBS Layer                                           & 82.00                & 65.00               & 95.69             & 88.43             & 82                   & 65                     & 67                   & 88                     \\
ResNet-50                                           & 81.82                & 70.25               & 95.53             & 91.31             & 82                   & 70                     & 71                   & 89                     \\
DenseNet-201                                        & 79.38                & 74.00               & 95.04             & 92.26             & 79                   & 74                     & 75                   & 91                     \\
NASNet                                              & 73.73                & 71.75               & 92.06             & 90.84             & 74                   & 72                     & 74                   & 86                     \\
\textbf{Emsemble}                                   & \textbf{83.42}       & \textbf{76.25}      & \textbf{96.31}    & \textbf{92.99}    & \textbf{83}          & \textbf{76}            & 73                   & 91                     \\
Lam et. al                                          & N/A                  & 57.2                & N/A               & N/A               & N/A                  & N/A                    & N/A                  & N/A                    \\ \bottomrule
\end{tabular}%
}
\end{table}

Results of rDR vs nrDR classification can also be found in Table \ref{tab:allResults}. All networks performed significantly better for this task than for normal vs abnormal classification on EyePACS dataset reaching maximum accuracy around 96\% with 99.44\% AUC using SBS layer architecture. For Messidor dataset, both NTS-Net and SBS Layer stand out from traditional classifiers. NTS-Net outperforms all other methods in all metrics, whereas ensemble of all methods gives sub-optimal performance than individual fine-grained methods. This can happen when majority of classifiers used for ensembling have a skewed performance towards downside and only a few give standout results.

\subsection{Results of Multi-Class Classification}
The complexity of classification task was gradually increased from binary to ternary and quaternary classification. Table \ref{tab:dist4c} and \ref{tab:dist3c} show the class distribution in train, validation and test splits for this multi-class setting. For ternary classification we used the criterion used by \cite{carson2018automated}, as shown in Figure \ref{fig:grades}.

\begin{table}
\parbox[t]{.49\linewidth}{
\centering
\caption{Class Distribution for 4-Class Classification}\label{tab:dist4c}
\resizebox{0.50\textwidth}{!}{ 
\begin{tabular}{ccccccc}
\hline
\multicolumn{1}{l}{\multirow{2}{*}{Grade}} & \multicolumn{3}{c}{EyePACS} & \multicolumn{3}{c}{Messidor}                                    \\ \cline{2-7} 
\multicolumn{1}{l}{}                       & Train  & Validate  & Test   & Train & \multicolumn{1}{l}{Validate} & \multicolumn{1}{l}{Test} \\ \hline
0                                          & 22668  & 2744      & 24741  & 346   & 49                           & 151                      \\
1                                          & 6157   & 1433      & 7196   & 107   & 16                           & 30                       \\
2                                          & 685    & 166       & 753    & 155   & 22                           & 70                       \\
3                                          & 490    & 126       & 733    & 92    & 13                           & 149                      \\
Total                                      & 30000  & 4469      & 33423  & 700   & 100                          & 400                      \\ \hline
\end{tabular}
}
}
\hfill
\parbox[t]{.49\linewidth}{
\centering
\caption{Class Distribution for 3-Class Classification}\label{tab:dist3c}
\resizebox{0.50\textwidth}{!}{ 
\begin{tabular}{ccccccc}
\hline
\multicolumn{1}{l}{\multirow{2}{*}{Grade}} & \multicolumn{3}{c}{EyePACS} & \multicolumn{3}{c}{Messidor}                                    \\ \cline{2-7} 
\multicolumn{1}{l}{}                       & Train  & Validate  & Test   & Train & \multicolumn{1}{l}{Validate} & \multicolumn{1}{l}{Test} \\ \hline
0                                          & 22668  & 2744      & 24741  & 346   & 49                           & 151                      \\
1                                          & 6157   & 1433      & 7196   & 107   & 16                           & 30                       \\
2                                          & 1175   & 292       & 1486   & 247   & 35                           & 219                      \\
Total                                      & 30000  & 4469      & 33423  & 700   & 100                          & 400                      \\ \hline
\end{tabular}
}
}
\end{table}

Performance of individual networks and their ensemble for ternary and quaternary classification is given in Table \ref{tab:allResults}. Ensemble of all models gave better performance in this case. We also observe that the performance of NTS-Net is higher than all other individual networks. Our accuracies for both ternary and quaternary classification are superior than accuracies reported by Lam et al. \cite{carson2018automated}. Figure \ref{fig:confMats} provides a detailed overview of classification performance of ensemble.

\begin{figure}[b!]
    \centering
    \addtocounter{figure}{-1}
    \begin{subfigure}{0.24\textwidth}
        \includegraphics[width=\textwidth]{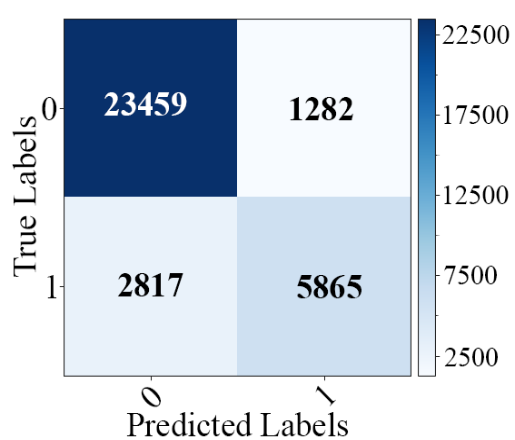}
        \caption{N/AB EyePACS}
        \label{fig:confMat-NAB-EP}
    \end{subfigure}
    \begin{subfigure}{0.24\textwidth}
        \includegraphics[width=\textwidth]{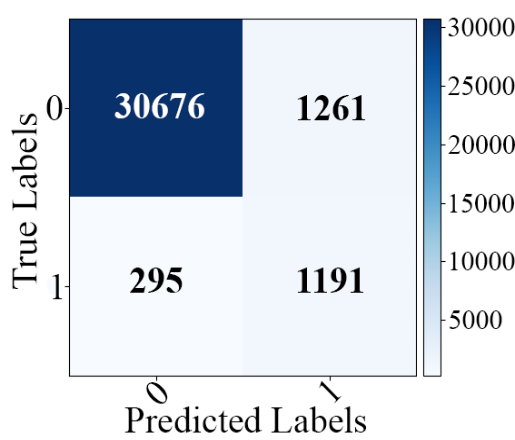}
        \caption{R/NR EyePACS}
        \label{fig:confMat-RNR-EP}
    \end{subfigure}
    \begin{subfigure}{0.24\textwidth}
        \includegraphics[width=\textwidth]{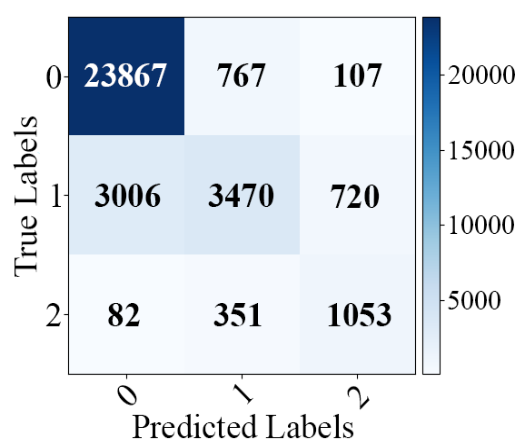}
        \caption{3-Class EyePACS}
        \label{fig:confMat-3c-EP}
    \end{subfigure}
    \begin{subfigure}{0.24\textwidth}
        \includegraphics[width=\textwidth]{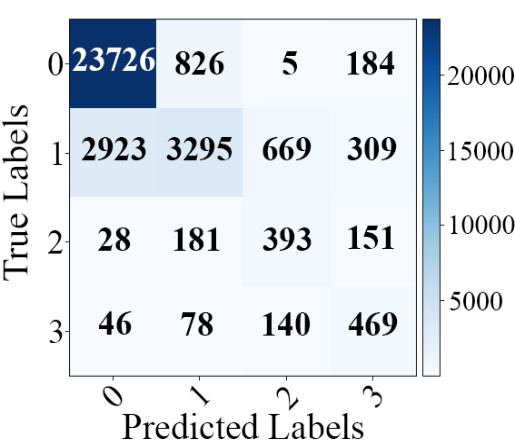}
        \caption{4-Class EyePACS}
        \label{fig:confMat-4c-EP}
    \end{subfigure}
    
    \begin{subfigure}{0.24\textwidth}
        \includegraphics[width=\textwidth]{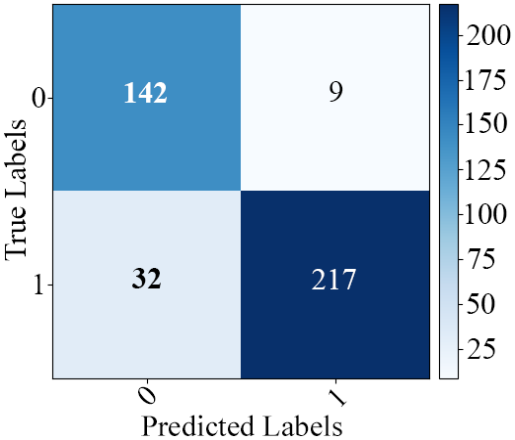}
        \caption{N/AB Messidor}
        \label{fig:confMat-NAB-M}
    \end{subfigure}
    \begin{subfigure}{0.24\textwidth}
        \includegraphics[width=\textwidth]{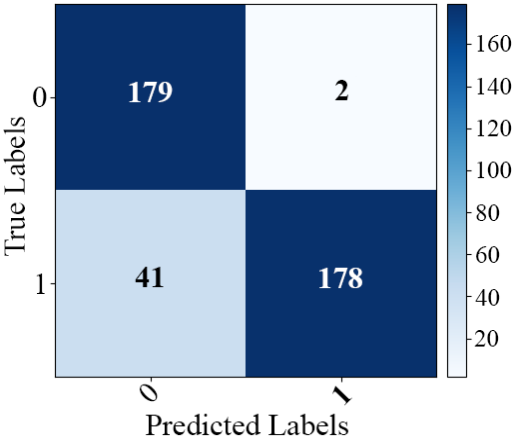}
        \caption{R/NR Messidor}
        \label{fig:confMat-RNR-M}
    \end{subfigure}
    \begin{subfigure}{0.24\textwidth}
        \includegraphics[width=\textwidth]{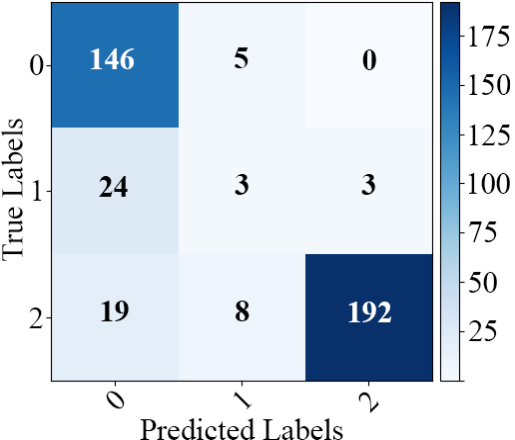}
        \caption{3-Class Messidor}
        \label{fig:confMat-3c-M}
    \end{subfigure}
    \begin{subfigure}{0.24\textwidth}
        \includegraphics[width=\textwidth]{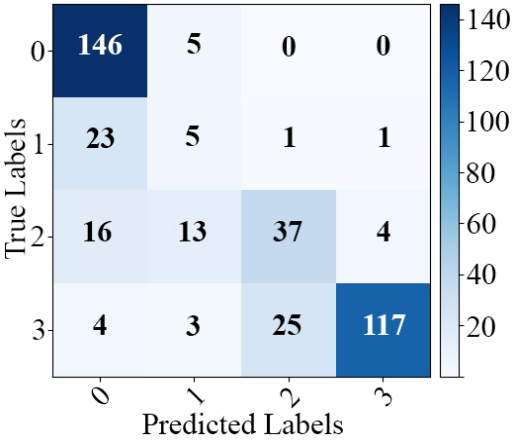}
        \caption{4-Class Messidor}
        \label{fig:confMat-4c-M}
    \end{subfigure}
    \caption{Confusion Matrices for EyePACS and Messidor for all Classification Tasks. N/AB refers to Normal vs Abnormal; R/NR refers to Referable vs Non-Referable}
    \label{fig:confMats}
\end{figure}

\section{Conclusion}
Diabetic Retinopathy detection using retinal fundus images is a fine-grained classification task. The biomarkers of this disease on retinal images are usually very small in size, especially for early stages, and are scattered all across the image. The ratio of pathologically important region to the whole input volume is therefore minuscule. Due to this reason traditional deep CNNs usually struggle to identify regions of interest and do not learn discriminatory features well. This problem of small and distributed visual artefacts coupled with unavailability of large publicly available high quality dataset with reasonable class imbalance makes DR detection particularly challenging for deep CNN models. However, fine-grained classification networks have high potential to provide standardized and large scale initial screening of diabetic retinopathy and help in prevention and better management of this disease. These networks are equipped with specialized algorithms to discover the important region from the image and pay particular heed to learn characterizing features from those regions.

We achieved superior performance for diabetic retinopathy detection on binary, ternary and quaternary classification tasks than many previously reported results. However, due to hugely different experimental setups and choice of performance metrics, it is unfair to draw a direct comparison with any of the cited research. Nevertheless, we have provided a wide spectrum of performance metrics and detailed experimental setup for comparison by any future work.

%
%
\bibliographystyle{splncs04}
\bibliography{references}
\end{document}